[a]sergei2@utk.edu

# Optimizing Training Trajectories in Variational Autoencoders via Latent Bayesian Optimization Approach

Arpan Biswas[1], Rama Vasudevan,[1] Maxim Ziatdinov[1,2] and Sergei V. Kalinin[3,a]

[1] Center for Nanophase Materials Sciences, Oak Ridge National Laboratory, Oak Ridge, TN 37831
[2] Computational Sciences and Engineering Division, Oak Ridge National Laboratory, Oak Ridge, TN 37831
[3] Materials Science and Engineering, University of Tennessee, Knoxville, TN 37996

Unsupervised and semi-supervised ML methods such as variational autoencoders (VAE) have become widely adopted across multiple areas of physics, chemistry, and materials sciences due to their capability in disentangling representations and ability to find latent manifolds for classification and/or regression of complex experimental data. Like other ML problems, VAEs require hyperparameter tuning, e.g., balancing the Kullback–Leibler (KL) and reconstruction terms. However, the training process and resulting manifold topology and connectivity depend not only on hyperparameters, but also their evolution during training. Because of the inefficiency of exhaustive search in a high-dimensional hyperparameter space for the expensive-to-train models, here we explored a latent Bayesian optimization (zBO) approach for the hyperparameter trajectory optimization for the unsupervised and semi-supervised ML and demonstrate for joint-VAE with rotational invariances. We demonstrate an application of this method for finding joint discrete and continuous rotationally invariant representations for MNIST and experimental data of a plasmonic nanoparticles material system. The performance of the proposed approach has been discussed extensively, where it allows for any high dimensional hyperparameter tuning or trajectory optimization of other ML models.

[a]sergei2@utk.edu

# 1. Introduction

Unsupervised and semi-supervised ML methods have become the mainstay of multiple domain areas ranging from machine vision to physics and astronomy due to their capability to disentangle representation and find latent manifolds for classification and/or regression tasks on complex raw data [1] [2], [3]. For sufficiently simple systems, the disentangled representations can often be associated with the specific physical factors of variability in the system. In particular, unsupervised ML approaches have allowed the discovery of physics from complex and/or large microscopic images/datasets as in [4]–[12], where the disentangled representations provide insight into specific physical order parameters.

As is common for unsupervised ML, the training of the model is sensitively dependent on the choice of hyperparameters. Generally, a hyperparameter is a parameter which controls the learning process of the ML models, and the hyperparameter tuning (or optimization) is the problem of choosing a set of optimal values for those hyperparameters to optimize the learning process. Extensive effort has been dedicated towards optimal tuning of ML models, with different optimization techniques such as gradient based method, genetic algorithm (GA), Bayesian optimization (BO) etc. [13]–[19]. In the process of tuning ML models where the training cost is computationally high, any exhaustive or manual parameter space search is a highly non-desirable approach. In such cases, BO is better suited than other optimization techniques due to the inbuild adaptive sampling towards maximizing the learning of region of interest within the parameter space while minimizing the function evaluation cost or training cost of expensive ML models. This approach has been widely used in these machine learning problems [20]–[24]. However, the downside of standard BO is the convergence issue when the control parameter is high dimensional (dimension >= 15-20) [25], resulting in a sub or non-optimal hyperparameter tuning of ML models. This again, will lead to improper ML training which ultimately results in poor physical insights from data. Additionally, the increase in the dimensionality of the control parameters decreases the rate (more function evaluations) of BO convergence, thus increasing the computational cost exponentially. This decreases the general applicability of using BO in optimizing an expensive ML model. Methods have been attempted to tackle BO in high dimensional problems through a different strategy of projection with random embedding and quantile Gaussian Process [26], [27] to a reduced space, or using special kernels [28]. However, performance of the method in ref [26] depends on the problem and the importance of parameters in the high-dimensional space whereas the method in ref [27] lacks computational efficiency. The method described in ref [28] builds on the technique on providing special attention to avoid excessive sampling over boundary region with a cylindrical kernel.

[a]sergei2@utk.edu

Here we attempt a different projection strategy of utilizing the variational autoencoder (VAE) model to incorporate the maximization of learning into the reduced latent dimension, and explore a Latent Bayesian Optimization (zBO) approach for high dimensional hyperparameter tuning or trajectory optimization without any prior knowledge on the potential optimality in the hyperparameter space, thereby lowering computational cost of expensive function evaluations and the risk of non-convergence of BO. Previously, similar approaches were attempted to solve for high dimensional and/or discrete input space as in [29]–[32].

However, here we design the zBO workflow for the high-dimensional continuous hyperparameter trajectory optimization, allowing for multiple independent high-dimensional input (trajectory function) space in a common reduced latent space, and particularly considering of the expensive, unsupervised joint rotationally invariant variational autoencoder (jrVAE) model where a trade-off between learning and function evaluation cost is critical without any prior knowledge from labeled data. This can be easily extended to any high dimensional hyperparameter (or parameter) optimization of other expensive ML (or black box) models. The overall integrated zBO-jrVAE framework is demonstrated on MNIST test problem and plasmonic nanoparticles material systems containing dataset of correlated scattering spectra of gold particles and SEM images.

The outline of this paper is as follows. Section 2 describes the general jrVAE and BO methods, and finally the proposed zBO algorithm, integrated to tune jrVAE model with a demonstration of the workflow on MNIST data. Section 3 demonstrates the application of zBO-jrVAE workflow plasmonic nanoparticles experimental dataset. Section 4 concludes the paper with final thoughts and potential future directions.

[a]sergei2@utk.edu

## 2. Methodology

In this section, we discuss on the key components focused for this paper, namely, Variational Autoencoder and Bayesian optimization. Finally, we discuss the Latent BO architecture implemented here, and its integration with the autoencoder model.

### 2.1. Joint Rotationally Invariant Variational Auto-Encoder (jrVAE)

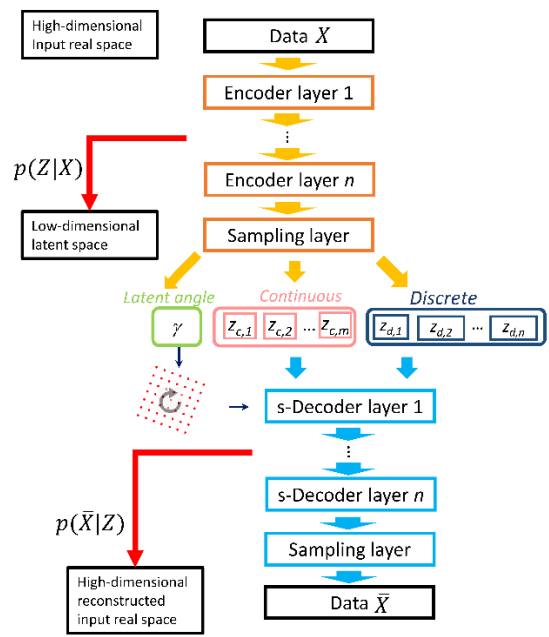

**Figure 1.** Joint rotational Variational Autoencoder (jrVAE) Framework.

A variational autoencoder is a deep generative probabilistic model that belongs to the family of probabilistic graphical models and variational Bayesian methods. The VAE is comprised of the encoder and decoder, as shown in Fig. 1. Given the input $x$, the encoder transforms it into a reduced latent space as a distribution, $p(z|x)$. Then, given any sample from the latent space $z$, the decoder reconstructs the input as $p(\bar{x}|z)$. The overall goal is to optimize the encoding-decoding process jointly to minimize the reconstruction error and the Kullback–Leibler divergence, to ensure the best representation of the latent space and maximize the restoration of the features in the input data. Here, the reconstruction error can be chosen as mean square error, the cross-entropy error, etc. The Kullback–Leibler divergence [33], [34] $D_{KL}(p(z|x)||p(z))$ is the distance loss

<a>asergei2@utk.edu</a>

between the prior $p(z)$ distribution (usually chosen as standard Gaussian) and the posterior $p(z|x)$ distributions of the latent representation from data. Different VAE models have been applied to materials systems, in attempt to learn from complex image data [35]–[39].

Here, we considered a specific example of joint rotationally invariant variational autoencoder model (jrVAE), as demonstrated in fig. 1, as implemented in pyroVED package in Python [40]. Here, the goal is to divide and learn the latent space into both continuous $p(z_c|x)$ and discrete $p(z_d|x)$ latent representation from the data, while enforcing rotational invariances. The loss function, $\mathcal{L}$, can be mathematically represented as follows:

$$\mathcal{L} = \varphi + \beta_c(i) D_{KL}(p(z_c|x)||p(z_c)) + \beta_d(i) D_{KL}(p(z_d|x)||p(z_d)) \tag{1}$$

where $\varphi$ is the reconstruction error, $D_{KL}(p(.|x)||p(.))$ is the Kullback–Leibler divergence, $\beta_c(i)$ and $\beta_d(i)$ are the continuous and discrete scale factors of KL divergence respectively at $i^{th}$ training cycle of jrVAE. The scale factors encourage a better disentanglement of the data, thus provides better learning [41]. However, as from eqn. (1), these are N-dimensional hyperparameters, where $d = e$, i.e, the dimension increases with the increase of training iteration $e$. Our objective in this paper is to tackle these high dimensional scale factors and build an optimization framework, balancing the accuracy in learning the optimal tuning and the computational cost of exploration.

## 2.2. Bayesian Optimization (BO)

Bayesian optimization (BO) [42], has been originally developed as a low computationally cost global optimization tool for design problems having expensive black-box objective functions. The general idea of BO is to emulate unknown functional behavior in the given parameter space and find the local and global optimal locations, while reducing the cost of function evaluations from expensive high-fidelity models. The reason it is called Bayesian is that it follows the tropes of Bayes theorem, which states that "posterior probability of a model (or parameters) M given evidence (or data, or observations) E is proportional to the likelihood of E given M, multiplied by the prior probability of M." Mathematically:

$$p(M|E) \propto \ell(E|M) p(M) \tag{2}$$

[a]sergei2@utk.edu

In the BO setting, the prior represents the belief of the unknown function $f$, assuming there exists a prior knowledge about the function (e.g., smoothness, noisy or noise-free etc.). Given the prior belief, the likelihood represents how likely is the observed data, $D$. These data are the sampled data as the true function value from expensive evaluations and can be viewed as the realizations from the unknown function. Finally, given the data, the posterior distribution is computed in BO as a posterior surrogate model, e.g. Gaussian process model (GPM): $\Delta = p(f|(D))$ is developed from these sampled data. Thus, eqn. 2 in the Bayesian optimization setting becomes:

$$p(f|(D_{1:k})) \propto \ell((D_{1:k})|f)p(f) \qquad (3)$$

where $D_{1:k} = [x_{1:k}, f(x_{1:k})]$ is the augmentation of the observation or sampled data till $k^{th}$ iteration of BO. Unlike general Bayesian modelling, here the prior or likelihood function is not written separately, but the augmentation of data at each iteration combines the prior distribution with the likelihood function. Generally, the GPM posterior model is considered based on conjugate normal prior assuming the data is realized from the Gaussian (normal) function.

Alongside with the major application of BO in problems with continuous response functions, attempts have been made when the response is discontinuous [43] or discrete such as in consumer modeling problems where the responses are in terms of user preference [42], [44]. Here, the user preference discrete response function is transformed into continuous latent functions using Binomial-Probit model for binary choices [45], [46] and polychotomous regression model is used for more than two choices where the user can state no preference [47]. However, the vast majority of these applications address for low-dimensional parameter optimization, especially when the function evaluations are expensive. Here, we aim to introduce a direction for BO application towards high-dimensional optimization problems with expensive evaluations.

[a]sergei2@utk.edu

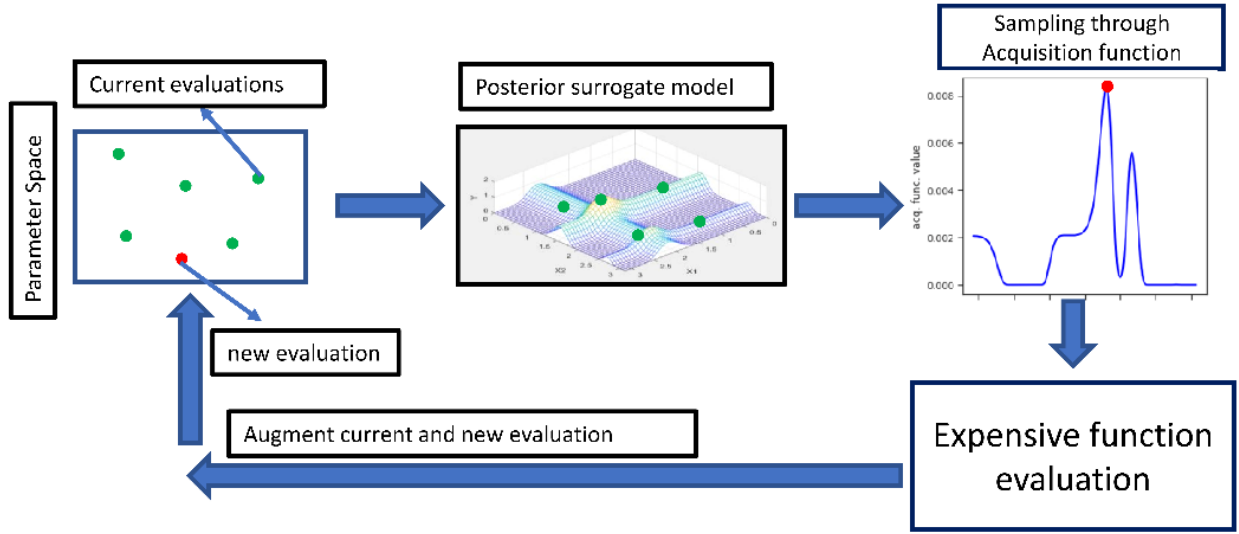

**Figure 2.** Bayesian Optimization (BO) Framework.

BO adopts a Bayesian perspective and assumes that there is a prior on the function; typically, a Gaussian process. The prior is represented from the true model or experiments which is assumed as the realizations of the true function, is treated as training data. The overall Bayesian Optimization Approach has two major components: A predictor or Gaussian Process Model (GPM) and an Acquisition Function (AF). As shown in fig. 2, in this approach, a posterior GPM is first built or iteratively updated, given the data from the current evaluations (training data). The surrogate GPM then predicts the realizations of the function at the unexplored locations over the control parameter space. The best locations are then strategically selected in the space for future expensive evaluations by maximizing the acquisition functions, defined from the posterior GPM simulations. Finally, after the evaluations are done, they are augmented with the existing evaluations and the cycle is repeated till the model is converged.

It is important to mention that as an alternative to a GPM, random forest regression has been proposed as an expressive and flexible surrogate model in the context of sequential model-based algorithm configuration [48]. Although random forests are good interpolators in the sense that they output good predictions in the neighborhood of training data, they are very poor extrapolators [49]. This can lead to selecting redundant exploration (more experiments) in the non-interesting region as suggested by the acquisition function in the early iterations of the optimization, due to having additional prediction error of the region far away from the training data. This motivates us to consider the GPM in a Bayesian framework while extending the application to high dimensional optimization problems.

[a]sergei2@utk.edu

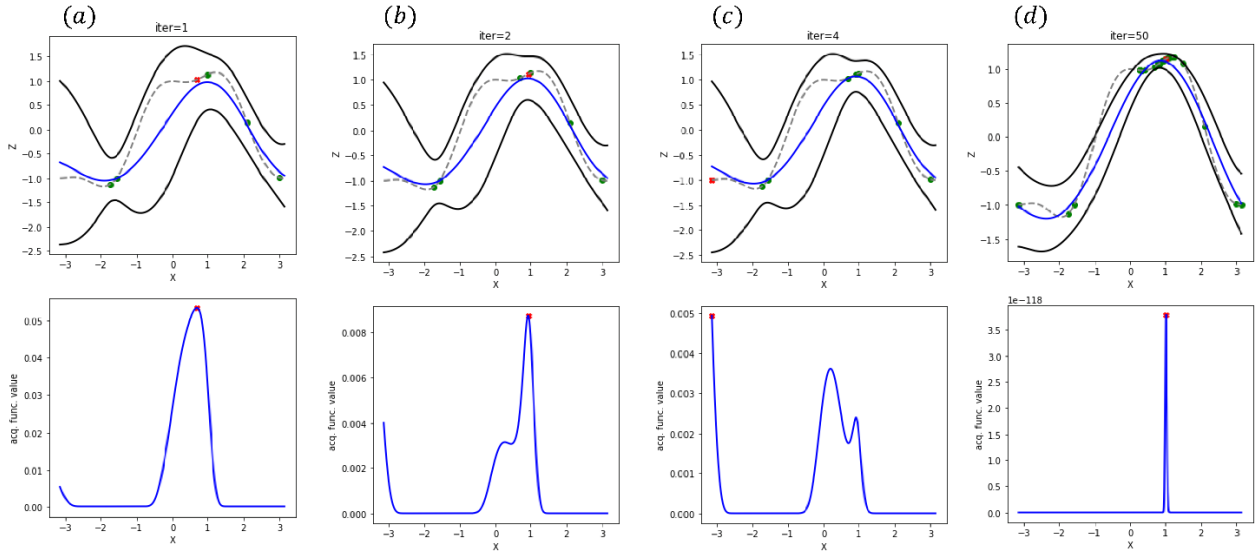

**Figure 3.** 1D Gaussian Process and search space exploration by maximizing acquisition function. The top images of (a-d) are the Gaussian process illustration. Gray dotted lines are the true function value, blue solid lines are the GP predicted mean, black solid lines are the GP mean ± 2x standard dev. Green dots are the current evaluated locations. Red dots are the new locations for next evaluations. The bottom images of (a-d) are the acquisition function illustration. Blue solid lines are acquisition function. Red dots are the new locations (at max. acq. func. value) for next evaluations.

As a simple example, fig. 3 (a-d) shows a simple 1D Gaussian Process Model with one control parameter $x$ and one objective function variable $z = f(x)$. The green dots are the evaluated locations, and the grey dotted and blue solid lines are the true and the predictor mean functions in the parameter space, given the prior evaluated data. The area enclosed by the black curves shows the measure of uncertainty over the surrogate GPM prediction. It is clearly seen that the variance near the observations is small and increases as the design samples are farther away from the evaluated data. Much research has been ongoing regarding incorporating and quantifying uncertainty of the experimental or training data by using a nugget term in the predictor GPM. It has been found that the nugget provides better solution and computational stability framework

[a]sergei2@utk.edu

[50], [51]. Furthermore, GPM has also been attempted in high dimensional design space exploration [52] and BIG DATA problems [53], as an attempt to increase computational efficiency. A survey of implementation of different GP packages has been provided in different coding languages such as MATLAB, R, and Python [54].

The general form of the GPM is as follows:

$$y(x) = x^T \beta + z(x) \tag{4}$$

where $x^T \beta$ is the Polynomial Regression model. The polynomial regression model captures the global trend of the data. In general, 1st order polynomial regression is used, which is also known as universal kriging [55]; however, it has also been claimed that it is fine to use a constant mean model [56]. $z(x)$ is a realization of a correlated Gaussian Process with mean $E[z(x)]$ and covariance $cov(x^i, x^j)$ functions defined as follows:

$$z(x) \sim GP\left(E[z(x)], cov(x^i, x^j)\right); \tag{5}$$

$$E[z(x)] = 0, cov(x^i, x^j) = \sigma^2 R(x^i, x^j) \tag{6}$$

$$R(x^i, x^j) = \exp\left(-0.5 * \sum_{m=1}^{d} \frac{\left(x_m^i - x_m^j\right)^2}{\theta_m^2}\right); \tag{7}$$

$$\theta_m = (\theta_1, \theta_2, \ldots, \theta_d)$$

where $\sigma^2$ is the overall variance parameter and $\theta_m$ is the correlation length scale parameter in dimension $m$ of $d$ dimension of $x$. These are termed as the hyper-parameters of GP model. $R(x^i, x^j)$ is the spatial correlation function. In this paper, we have considered a Radial Basis function which is given by eqn. 7. The objective is to estimate (by MLE) the hyper-parameters $\sigma$, $\theta_m$ which creates the surrogate model that best explains the training data $D_k$ at iteration $k$.

After the GP model is fitted, the next task of the GP model is to predict at an arbitrary (unexplored) location drawn from the parameter space. Assume $D_k = \{X_k, Y(X_k)\}$ is the prior information from previous evaluations or experiments from high fidelity models, and $\bar{\bar{x}}_{k+1} \in \bar{\bar{X}}$ is a new design within the unexplored locations in the parameter space, $\bar{\bar{X}}$. The predictive output distribution of $x_{k+1}$, given the posterior GP model, is given by eqn 8.

[a]sergei2@utk.edu

$$P(\bar{\bar{y}}_{k+1}|D_k, \bar{\bar{x}}_{k+1}, \sigma_k^2, \boldsymbol{\theta}_k) = N(\mu(\bar{\bar{y}}_{k+1}(\bar{\bar{x}}_{k+1})), \sigma^2(\bar{\bar{y}}_{k+1}(\bar{\bar{x}}_{k+1}))) \qquad (8)$$

where:

$$\mu(\bar{\bar{y}}_{k+1}(\bar{\bar{x}}_{k+1})) = cov_{k+1}^T COV_k^{-1} Y_k; \qquad (9)$$

$$\sigma^2(\bar{\bar{y}}_{k+1}(\bar{\bar{x}}_{k+1})) = cov(\bar{\bar{x}}_{k+1}, \bar{\bar{x}}_{k+1}) - cov_{k+1}^T COV_k^{-1} cov_{k+1} \qquad (10)$$

$COV_k$ is the kernel matrix of already sampled designs $X_k$ and $cov_{k+1}$ is the covariance function of new design $\bar{\bar{x}}_{k+1}$ which is defined as follows:

$$COV_k = \begin{bmatrix} cov(x_1, x_1) & \cdots & cov(x_1, x_k) \\ \vdots & \ddots & \vdots \\ cov(x_k, x_1) & \cdots & cov(x_k, x_k) \end{bmatrix}$$

$$cov_{k+1} = [cov(\bar{\bar{x}}_{k+1}, x_1), cov(\bar{\bar{x}}_{k+1}, x_2), \ldots, cov(\bar{\bar{x}}_{k+1}, x_k)]$$

The second major component in Bayesian optimization is the Acquisition Function (AF). AF guides the search for future evaluations or experiments towards the desired goal and thereby bring the sequential design into the BO. The AF predicts an improvement metric for each sample. The improvement metric depends on exploration (probability to discover useful behaviors in unexplored locations) and exploitation (known region with high responses). Thus, the acquisition function gives high value of improvement to the samples whose mean prediction is high, variance is high, or a combination of both. By maximizing the acquisition function, we select the best samples to find the optimum solution and reduce the uncertainty of the unknown expensive design space.

Fig. 3 (a-d) shows a simple example of BO exploration with one control parameter *x* and one objective function variable $z = f(x)$ of the sequential selection of samples by maximizing the acquisition function, given posterior GP model in iterations 1, 2, 4 and 50. The solid blue line is the GP predicted function and the dotted grey line is the original function. The green dots in iteration 1 (3a. top figure) are the evaluated locations and the red dot is the new evaluated location, as per maximizing the acquisition function (3a. bottom figure). The new data is augmented with current data and the GPM is updated in the next iteration, and then the acquisition function is maximized to get the next evaluated locations (see fig. 3b). This process is repeated iteratively, and in this manner the search space is explored. We can see from the figure that the acquisition function value is highest where the samples have high prediction mean and/or high variance and the lowest where the samples have low prediction, low variance or both. Thus, the acquisition function strategically selects points which have the likelihood to have the optimal (e.g. maximum value of the unknown function) and gradually reduces the error through sequential (Bayesian) learning towards aligning with the true function, at the region of interest (see fig. 3d).

[a]sergei2@utk.edu

Throughout the years, various formulations have been applied to define the acquisition functions. One such method is the Probability of Improvement, PI [57] which is improvement-based acquisition function. Jones [58] notes that the performance of PI(·) "is truly impressive;… however, the difficulty is that the PI(·) method is extremely sensitive to the choice of the target. If the desired improvement is too small, the search will be highly local and will only move on to search globally after searching nearly exhaustively around the current best point. On the other hand, if the small-valued tolerance parameter ξ in PI(.) equation is set too high, the search will be excessively global, and the algorithm will be slow to fine-tune any promising solutions." Thus, the Expected Improvement acquisition function, EI [42], is widely used over PI which is a trade-off between exploration and exploitation. Another Acquisition function is the Confidence bound criteria, CB, introduced by Cox and John [59], where the selection of points is based on the upper or lower confidence bound of the predicted design surface for maximization or minimization problem respectively.

### 2.3. Latent Bayesian Optimization (zBO): Integrated to jrVAE

As our objective mentioned in section 2.1, to optimize the high dimensional KL factors or KL trajectory, we considered BO technique due to adaptive sampling (from acquisition function) for fast learning and find the region of interest in the parameter space. However, the standard BO (Section 2.2) is not well suited to tackle high-dimensional parameter optimization. Hence, here we illustrate the modification to build a Latent Bayesian Optimization (zBO). Figure 4 shows the overall workflow of zBO, integrated to jrVAE model. Table 1 shows the detailed algorithm of the workflow. However, the zBO framework is a standalone application, which can be easily coupled with other ML or mathematical models, which requires high dimensional (hyper)parameter optimization.

[a]sergei2@utk.edu

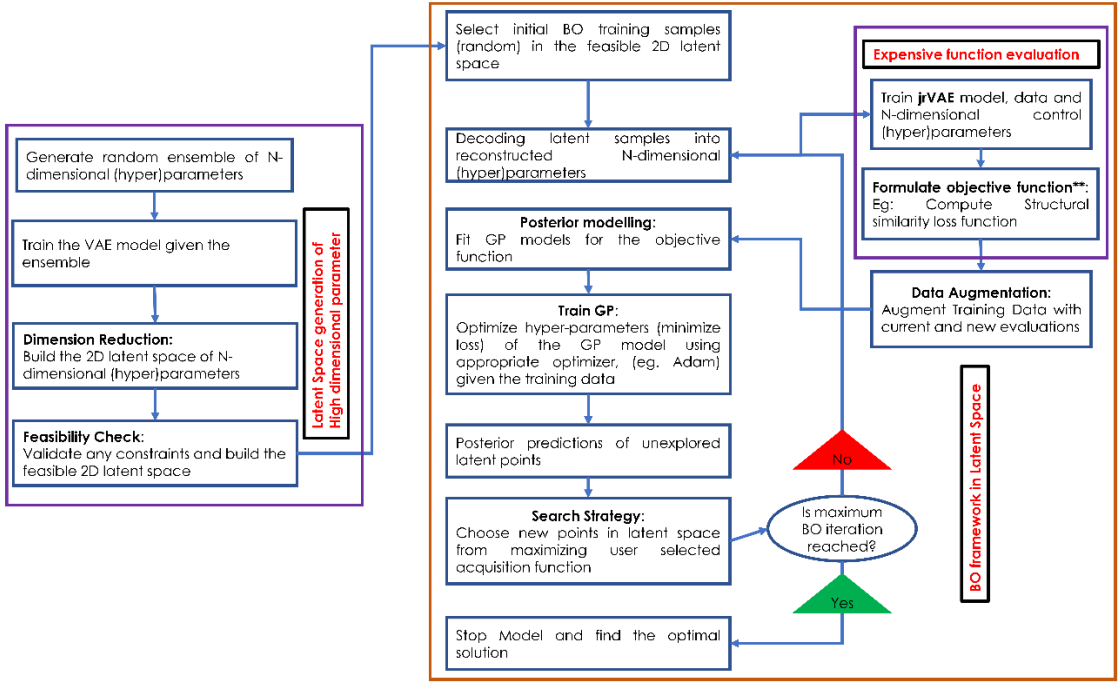

**Figure 4.** Latent Bayesian Optimization (zBO) architecture: Application to high-dimensional hyperparameter tuning of jrVAE model for performance enhancement.

**\*\*Note**: The objective function formulation is based on the analysis on this paper, and can be easily updated in the framework with different formulation as per the problem requirement.

*Table 1: Algorithm: Latent Bayesian Optimization: To optimize high-dimensional hyperparameter tuning of jrVAE*

1. **Initialization of high-dimensional parameters**: Generate, either random or with some prior knowledge, a set of N-dimensional scale ($\beta$) trajectories. N represents the number of training cycles of jrVAE model.

2. **Training VAE:** Define and train a standard VAE model on the set of $\beta$ trajectories. We assume here the VAE model is tuned properly for maximized learning.

3. **Dimension Reduction of high-dimensional parameters:** With the trained VAE model (step 2), we build a 2D latent space of N-dimensional $\beta$ trajectories.

4. **Feasibility Check:** Formulate all the physical constraints (if any). We validate the 2D latent space for any constraint's violations. In this case, $\beta(i) > 0; i = 1, 2, .., N$. Finally, build the feasible 2D latent space**.**

5. **Initialization for BO:** State maximum BO iteration, $M$. Randomly select $j$ samples from the feasible 2D latent space, $Z = \{Z_1, Z_2\}$. Assuming $f$ is the expensive objective function. Set $k = 1$.

[a]sergei2@utk.edu

For $k \leq M$

6. **Decoding into reconstructed high-dimensional parameters for expensive function evaluations**: Given the trained VAE model (Step 2), decode each latent sample $z = \{z_1, z_2\}$; $z \in Z_k$ into reconstructed N-dimensional sampled $\beta$ trajectories, as $\bar{x}|z = \{\bar{x}_1, \bar{x}_2, \ldots \bar{x}_N | z_1, z_2\}$; $\bar{x} \in \bar{X}_k$. Evaluate $j$ samples for objective as, $y_j(\bar{x}|z); y \in Y_k$. The detailed formulation of the objective function in this case is provided later in the section. Build training data matrices (in 2D latent space), $D_k = \{Z_k, Y_k\}$.

7. **Surrogate Modelling**: Develop or update GPM models, given the training data, as $\Delta(D_k)$.
   a. Optimize the hyper-parameters of GPM by minimizing the loss (negative marginal log-likelihood) function using Adam optimizer algorithm. Here, we consider learning rate 1e-4.

8. **Posterior Predictions**: Given the surrogate model, compute posterior means and variances for the unexplored locations, $\overline{\overline{Z_k}}$, over the 2D latent space as $\pi(Y(\overline{\overline{Z_k}})|\Delta$ and $\sigma^2(Y(\overline{\overline{Z_k}})|\Delta$ respectively. It is to be noted that we directly compute the posterior predictions from the input 2D latent samples, given the GPM, without the need to decode to high dimensional parameter.

9. **Acquisition function:** Compute and maximize acquisition function, $\max_z U(f|\Delta)$ to select next best location in the 2D latent space, $z_{best,k}$ for evaluations.

10. **Augmentation:** Following step 6, decode $z_{best,k}$ into $\bar{x}_{best,k}|z_{best,k}$, and evaluate the same as $y(\bar{x}_{best,k}|z_{best,k})$. Augment data, $D_{k+1} = [D_k; \{z_{best,k}, y\}]$.

We further elaborate Step 6 of the stated Algorithm (Table 1) and describe the workflow of the objective function considered in this paper. However, to generalize, the proposed approach can be easily coupled with any objective functions as required for a given problem. In this paper, our general task is to undergo different multi-label classification problems through unsupervised learning techniques of jrVAE model, and the attempt for better quality of solution from optimal tuning of scale ($\beta$) trajectories through zBO framework. In the experimental data, often we don't possessany prior knowledge of discrete classes for supervised or semi-supervised learning, thus, we have focused on the unsupervised ML learning as the problem domain in this paper. Table 2 provides the workflow of the objective function evaluation, which aids the task of better separation of different classes from the data.

14[a]sergei2@utk.edu

*Table 2: Workflow: Objective Function evaluation (for problem of multi-label classification) as in Step 6 and 10 of Table 1*

1. **Train jrVAE model**: Initialize number of discrete classes, $D_c$. Given the data (simulated or experimental), and the decoded $\beta$ trajectories, as $\bar{x}|z$, train jrVAE model for N training cycles. Compute the 2D trained manifold for each discrete classes, as in matrix $\boldsymbol{\Phi}$.

2. **Evaluate objective 1:** Choose learned manifolds $\boldsymbol{\Phi_i}, \boldsymbol{\Phi_j}$ for two discrete classes, $i, j; j > i; i, j = 1, 2, \ldots D_C$. Calculate structural similarity (SSIM) loss function between them. Do the same between every discrete class. Compute the total loss as $\ell_1(\boldsymbol{\beta}|data) = \sum_{i,j;j>i}^{D_c} \ell_{i,j}$.

3. **Evaluate objective 2:** Choose learned manifold $\boldsymbol{\Phi_i}$ for a discrete class $j; j = 1, 2, \ldots D_C$. Randomly choose unique $I$ pairs, as $i_1, i_2; i_1 \neq i_2; i_1, i_2 = 1, 2, \ldots I$, of image grid location within the learned manifold. Calculate structural similarity (SSIM) loss function between each pair. Compute the mean loss for learned manifolds of each discrete class as $\ell_j = \frac{\sum_{i_1,i_2;i_1 \neq i_2}^{I} \ell_{i_1,i_2}}{k}$ where $k$ is the number of unique location pairs. Do the same for every discrete class. Compute the total loss as $\ell_2(\boldsymbol{\beta}|data) = \sum_j^{D_c} \ell_j$. We avoid computing for every possible combination of grid location for significant (exponential) increase in computational cost as the number of grid locations and the discrete classes increases.

4. **Formulate final objective function:** Here our goal is to maximize the SSIM loss of learned manifolds among different discrete classes, thus maximize objective 1, to minimize the SSIM loss within the images of the grid location of a learned manifold of a discrete class, thus minimize objective 2. Finally, to modify into a maximization problem (as per default setting of zBO), the objective function is stated as

$$\max_{\boldsymbol{\beta}} \ell_1(\boldsymbol{\beta}|data) - \ell_2(\boldsymbol{\beta}|data) \tag{11}$$

To demonstrate the proposed workflow, we first considered a test MNIST dataset [60], which is a large database of handwritten digits, commonly used for training various image processing systems. The database contains 60000 training datasets, where each digit can be considered as a label for our multi-class classification problem. To note, the digits contain rotational variability, which is a target to tackle by jrVAE model, and with the hyperparameter tuning.

In this case study, we considered gaussian decoder sampler with sigma = 0.3, learning rate = 1e-4 and training cycles =1000 to initialize and train the VAE model for 2D latent representation of N-dimensional $\beta$ trajectories (Table 1, Step 2). For initializing the jrVAE model, we considered Bernoulli decoder sampler, learning rate = 1e-3 and training cycle, $N = 120$. For the zBO, we started with 20 randomly selected samples with maximum of 120 BO iteration, thus a total of 140 function evaluations. We choose Expected Improvement (EI) acquisition function. To simplify the problem, we considered to optimize the continuous scale factor, $\boldsymbol{\beta}_c$ trajectory only and set the discrete scale factor at constant setting as $\beta_d(i) = 3; i = 1, 2, ..., N$ where $N = 120$ in this case. However, the proposed workflow can be easily extended to optimize both scale factors jointly, which is considered in future scope.

[a]sergei2@utk.edu

During the function evaluation process within BO, we considered a subset of data (randomly selected) to avoid redundant computational cost from training with a large dataset (as we need function evaluations multiple times). Since the dataset is balanced, it is easy to work with the subset of the data and the objective function should be able to still capture the necessary information from data to find the optimal region. Also, we avoided the loss function computation of Table 2, Step 3 as we observe the other loss function is sufficient to achieve the optimal region for this case study, and therefore further reducing the computational expense. Thus, in this case study, we only consider the first part of eqn. 11. the Once the zBO model is converged, we train the jrVAE model with full (large) dataset, stated settings, and optimal tuning of $\boldsymbol{\beta}_c$.

Figure 5 shows an example of the projection of N-dimensional search space to the feasible 2D latent search space, following the algorithm in Table 1, Step 1-4. To increase the complexity, we defined the training trajectories from two different functions: 1) linear cooldown and 2) randomly segmented with random noise (see fig. 5a). We modified the training data at the same scale. With VAE training and constraint validation, we defined the feasible 2D latent space (fig. 5b) where each latent samples can be reconstructed to a $\boldsymbol{\beta}_c$ trajectory (fig. 5c-e). The blue dots over the 2D latent space are the training data. We can clearly see the trained VAE model build clusters of two different pattern of trajectories (defined from separate functions) and the decoding provides the pattern of reconstructed trajectories (red dots) with a weighted information, depending on the distance from these clusters (comparing figs 5c, d and e). This shows the VAE model is well trained which not only restore sufficient knowledge of each defined trajectories separately (storing original patterns) but also provides reconstruction with mixing of both knowledge

[a]sergei2@utk.edu

(introducing new hybrid patterns). This increases the possible set of new solutions for zBO without losing pre-considered solutions.

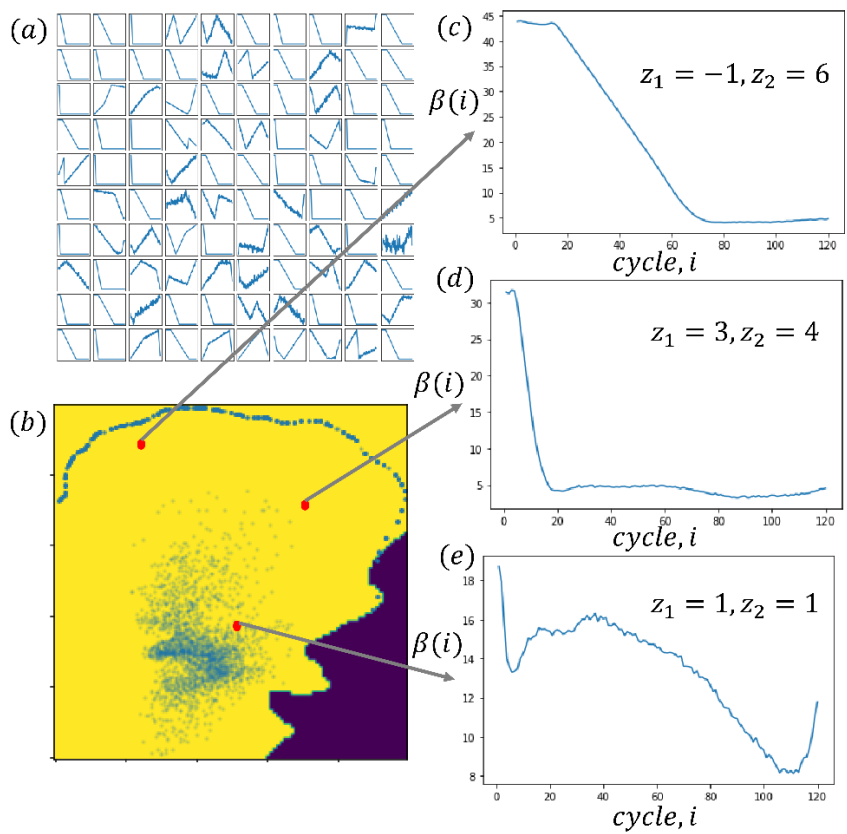

**Figure 5.** Workflow to generate a feasible 2D latent search space from N-dimensional (hyper)parameter space. (a) Samples of different $\boldsymbol{\beta}_c$ trajectories (here defined from two different functions) as input for training VAE model (ref Table 1, step 2). (b) 2D latent representation of the $\boldsymbol{\beta}_c$ trajectories, given the trained VAE model. The blue dots are the training data as shown in (a). The light area and dark area are the feasible and infeasible region. The feasible region is only considered for the search space during optimization. (c), (d) and (e) are the reconstructed $\boldsymbol{\beta}_c$ trajectories from the trained VAE decoder at locations (red dots): $(z_1, z_2) = (-1, 6)$, $(z_1, z_2) = (3, 4)$ and $(z_1, z_2) = (1, 1)$ respectively.

Figure 6 shows the result of zBO convergence, and the optimal $\boldsymbol{\beta}_c$ trajectory, whereas Figure 7 shows the comparison among different VAE models, with and without the high-dimensional continuous scale factor tuning. It is clear that the vanilla VAE model provides the worst result (see fig. 7b), where it could not extract all the labels from data and has rotational variability in the learned manifolds. With the jrVAE model, but with default constant setting of $\beta_c(i) = 1; i = 1, 2, \ldots, 120$, though the model improved much and could be able to separate the

ᵃsergei2@utk.edu

labels with rotational invariances, there are still some mixings of labels at some locations of the trained manifolds (see fig. 7c). We see once the jrVAE trained with optimal tuning of $\boldsymbol{\beta_c}$, we get the best solution among other scenarios, where all discrete classes separated efficiently with negligible mixing of labels at any locations of the trained manifolds (see 7d).

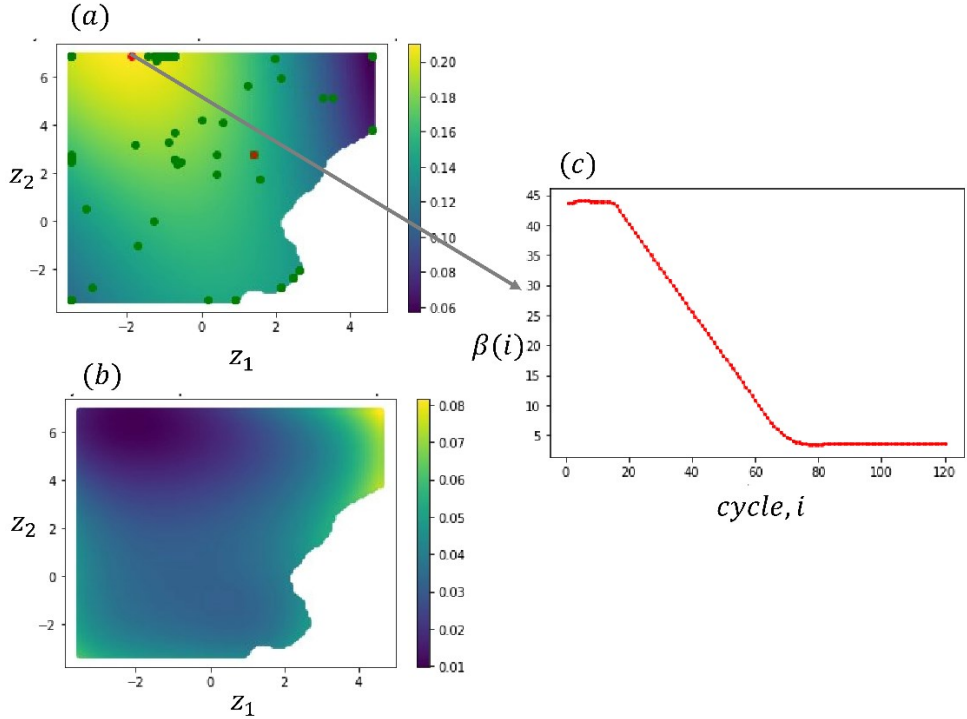

**Figure 6.** Results of zBO to find optimal region of $\boldsymbol{\beta_c}$ trajectories. (a), (b) GP estimated objective function (SSIM loss) map and its uncertainty map in the latent space respectively. The red dot is the maximum estimated objective value, and the red cross is the maximum among evaluated samples. We consider the GP estimated maximum value as the optimal solution and (c) is the respective reconstructed $\boldsymbol{\beta_c}$ trajectory at that optimal location (red dot) in the feasible 2D latent space.

[a]sergei2@utk.edu

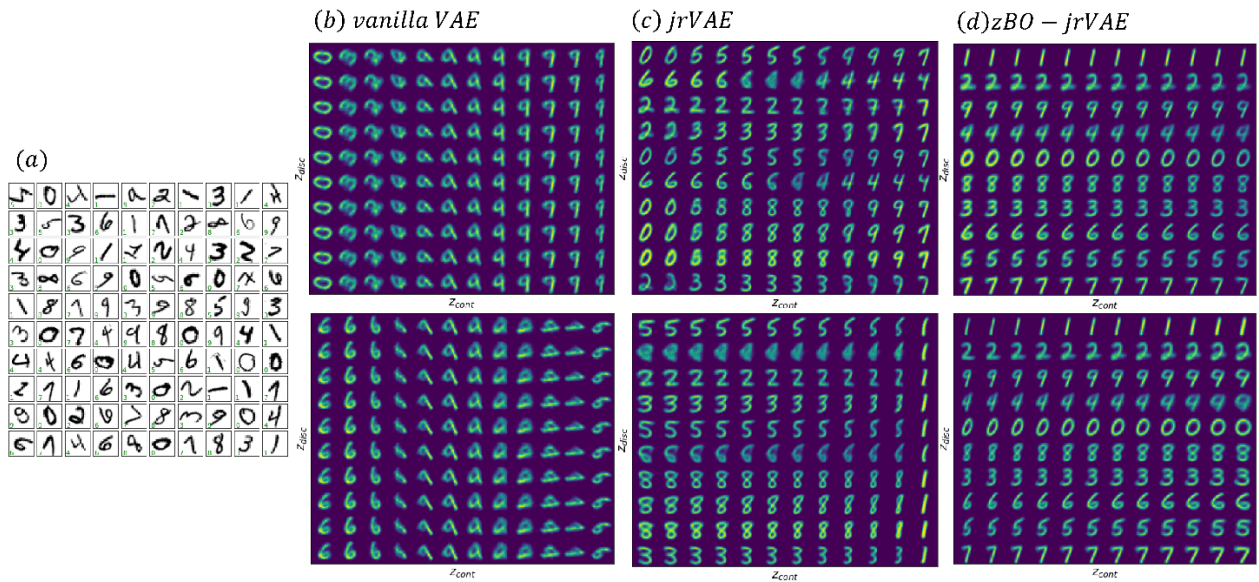

**Figure 7.** Comparison of different VAE models. (a) Few training samples of MNIST dataset as the task of multi-label (digits) classification problem. We have trained different VAE model for 120 cycles, given the full MNIST data (60000 training data). The learned manifolds traversal for the 2D latent space (two latent variables) as shown in rows after training of respective (b) Vanilla VAE model, (c) jrVAE model, with default setting of $\boldsymbol{\beta}_c$ trajectory (no zBO implementation) and (d) jrVAE model, with optimal tuning (as shown in fig. 6c) of $\boldsymbol{\beta}_c$ trajectory (zBO implementation).

## 3. Results

In this section, we showcased the proposed zBO-jrVAE workflow to an experimental data of plasmonic nanoparticles. The datasets contain correlated scattering spectra of gold particles and scanning electron microscope images. Here, our task is to conduct the multi-label classification, where we attempt to extract and separate the discrete labels as particle counts in the images through the unsupervised learning of jrVAE model, integrated with zBO.

In this case study, we considered gaussian decoder sampler with sigma = 0.3, learning rate = 1e-4 and training cycles =2000 to initialize and train the VAE model for 2D latent representation of N-dimensional $\beta$ trajectories (Table 1, Step 2). For initializing the jrVAE model, we considered gaussian decoder sampler with sigma = 0.01, learning rate = 1e-4 and training cycle, $N = 200$. It is to be noted that here we have different input scale factor dimension ($N = 200$) unlike the MNIST analysis ($N = 120$), depending on the number of training cycles of jrVAE model. However, the dimension of the problem in zBO (during optimization) can be still represented as

[a]sergei2@utk.edu

2D (in latent space), and does not increase with the increase of dimension of the input in real space. For the zBO, we started with 20 randomly selected samples with maximum of 100 BO iteration, thus a total of 120 function evaluations. We choose Expected Improvement (EI) acquisition function. Here also, we optimize the continuous scale factor, $\beta_c$ trajectory only and set the discrete scale factor at constant setting as $\beta_d(i) = 0.01; i = 1, 2, ..., N$ where $N = 200$ in this case. We first normalize and balanced (through weighted resampling technique) the raw experimental data set. During the function evaluation process within BO, as previous analysis, we considered a subset of data (randomly selected). We executed the zBO on NVIDIA's DGX-2 server. Once the zBO model is converged, we train the jrVAE model with large dataset, stated settings, and optimal tuning of $\beta_c$. We use the Google Colab Pro to execute the final training. However, unlike the previous analysis, here we considered the full loss function (Step 2, 3 of Table 2) as per eqn. 11.

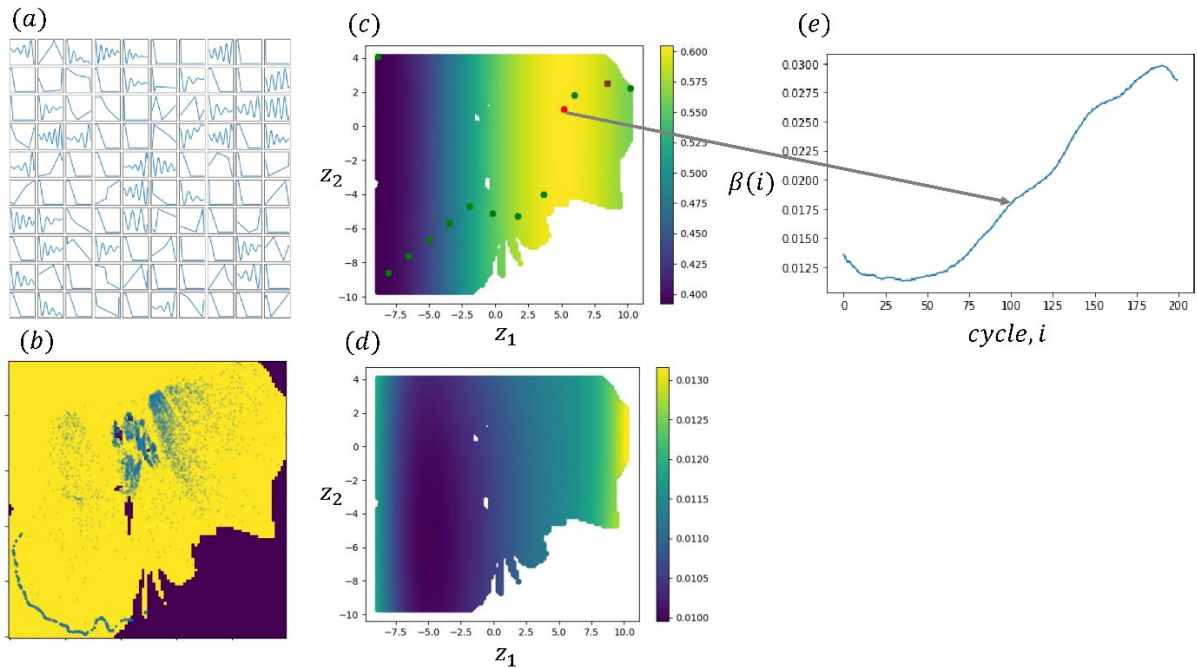

**Figure 8.** Classification problem of Plasmonic Nanoparticles data (3 discrete labels). Results of zBO to find optimal region of $\beta_c$ trajectories. (a) Samples of different $\beta_c$ trajectories (here defined from three different functions) as input for training VAE model. (b) 2D latent representation of the $\beta_c$ trajectories where the blue dots are the training data (c), (d) GP estimated objective function (SSIM loss) map and its uncertainty map in the latent space respectively. The red dot is the maximum estimated objective value, and the red cross is the maximum among evaluated samples. We consider the GP estimated maximum value as the optimal solution and (e) is the respective reconstructed $\beta_c$ trajectory at that optimal location (red dot) in the feasible 2D latent space.

[a]sergei2@utk.edu

We started with the classification of three labels as images with one, two and three particles. Fig. 8 shows the projection of N-dimensional search space to the feasible 2D latent search space. In this case, we defined the training trajectories, with same scale, from three different functions: 1) linear cooldown, 2) randomly segmented and 3) periodic (see fig. 8a). Fig. 8b shows the respective feasible 2D latent representation. Fig. 8c-e shows the result of zBO convergence, and the optimal $\beta_c$ trajectory. The model converged earlier than 100 function evaluations as the acquisition function value goes to negligible, meaning the optimal region is already found with negligible uncertainty and no further meaningful learning is possible as trade-off with expensive function evaluations at any other locations. We can see the optimal trajectory is very different than what we found in solving MNIST problem (see fig 6c).

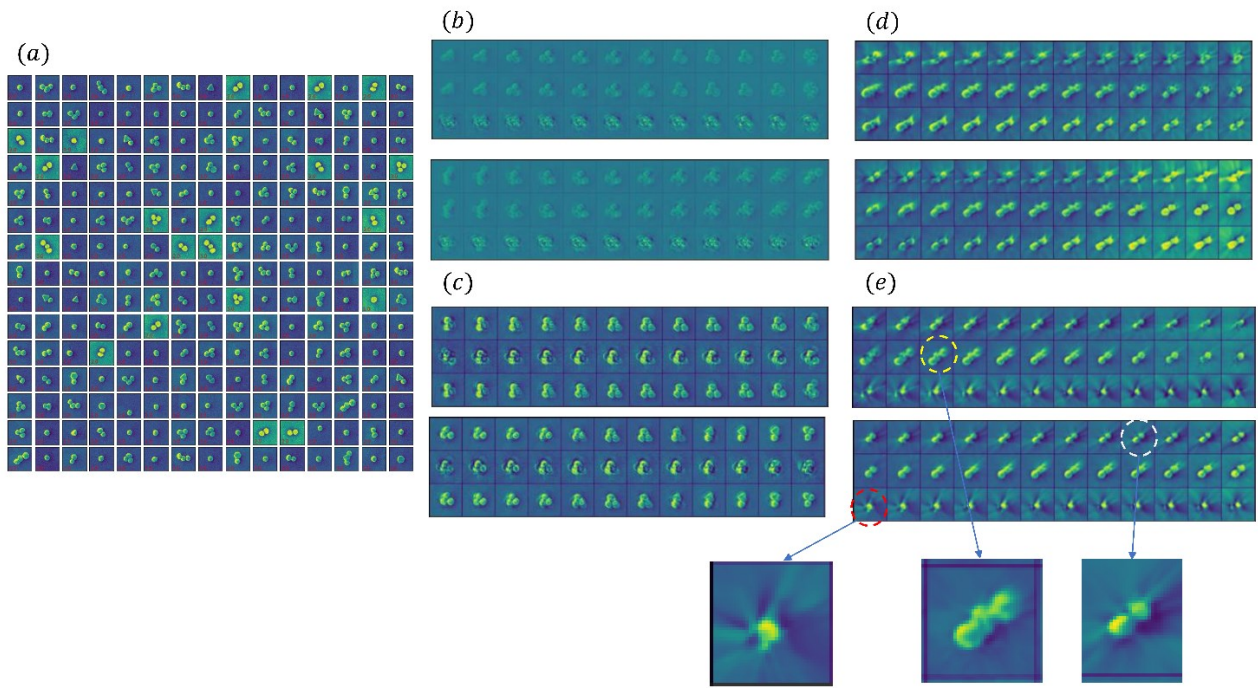

**Figure 9.** Comparison of different VAE models. (a) Few training samples of plasmonic nanoparticles dataset, considering 3 discrete labels (particle counts). We have trained different VAE model for 200 cycles. The learned manifolds traversal for the 2D latent space (two latent variables) as shown in rows after training of respective (b) Vanilla VAE model (all parameter in default settings as in [40]), (c) VAE model with other hyperparameter setting as stated for this case study (d) jrVAE model, with default setting of $\beta_c$ trajectory (no zBO implementation) and (e) jrVAE model, with optimal tuning (as shown in fig. 8c) of $\beta_c$ trajectory (zBO implementation).

[a]sergei2@utk.edu

Fig. 9 shows the comparison among different VAE models, with and without the high-dimensional continuous scale factor tuning. We can clear see the vanilla VAE model provides the worst result again (see fig. 9b). Even with tweaking other hyperparameters, we able to get the images with distinct features (see fig. 9c), however for both cases the models could not separate the labels and thus provide no physical insight from the data. With the jrVAE model, but with default constant setting of $\beta_c(i) = 1; i = 1, 2, ..., 200$, though the model shows some separation of classes (class of two particles), there are still some mixings of labels specially at top rows of the trained manifolds (see fig. 9d), which results into some infeasible physical behavior. We see once the jrVAE trained with optimal tuning of $\boldsymbol{\beta_c}$, we see further enhancement and get the best solution among other scenarios. As in fig. 9e, we see the top and the bottom rows of the learned manifolds have distinct classes of two and one particles respectively, whereas the middle row contains classes of two and three particles at different locations. We have also compared the results (in Supplementary material, fig. A1.) with optimal tuning of $\boldsymbol{\beta_c}$ found in solving MNIST problem (fig. 6c) to understand the sensitivity of $\boldsymbol{\beta_c}$ for a given problem. We see the optimal setting found for the nanoparticles problem provides better training of jrVAE, in better separating into discrete class (rows) of the manifolds.

Similar analysis to materials and methods is done utilizing the GPU server for fast training, with more complex nanoparticles dataset where we considered 7 labels (particles counts). A sample of the dataset is provided in Supplementary material as fig. A2. Workflow on 2D latent representation and optimizing $\boldsymbol{\beta_c}$ through zBO is illustrated in fig. 10. Here also, the model converged earlier than 100 function evaluations as the acquisition function value goes to negligible. Fig. 11 shows the similar comparative analysis among different VAE models. Here also, the Vanilla VAE models (figs. 11a, 11b) not able to separate any labels and therefore fail to extract any valuable physics from data. Surprisingly, the jrVAE model with default constant setting of $\beta_c(i) = 1; i = 1, 2, ..., 200$ gives very poor solution with too distorted images to identify any labels properly, which results into an unrealistic extraction of physical information (see fig 11c.). Similarly with optimal tuning of $\boldsymbol{\beta_c}$, shows the best disentanglement of data among other scenarios (see fig 11d.). Though we could be able to extract 5 discrete labels (particle counts 1 to 5) out 7, comparatively for the rest of the scenarios, we have not achieved any valuable physical information. This shows a much better improvement, where we able to extract $> 70 \%$ of physical insights, with the optimization approach through zBO. We believe for further improvement may need the optimization of $\boldsymbol{\beta_d}$, which is a task for future investigation. However, it is evident that the extraction of knowledge from experimental data is much harder than test data (like in MNIST), however, we still get a very good improvement from optimal tuning with zBO in learning plasmonic nanoparticle system, where we can get an appropriate physical insight of all the classes from the data (considering 3 labels) and above 70% physical information from the data (considering 7 labels). The purpose of this work is the zBO framework to guide towards finding the maximum (not necessarily always 100 %) learning of physical insights from optimizing high dimensional parameters or hyperparameters of other mathematical or ML models for a given application, given the degree of complexity (variability) of the problem (data) and the fixed settings of other parameters or hyperparameters.

[a]sergei2@utk.edu

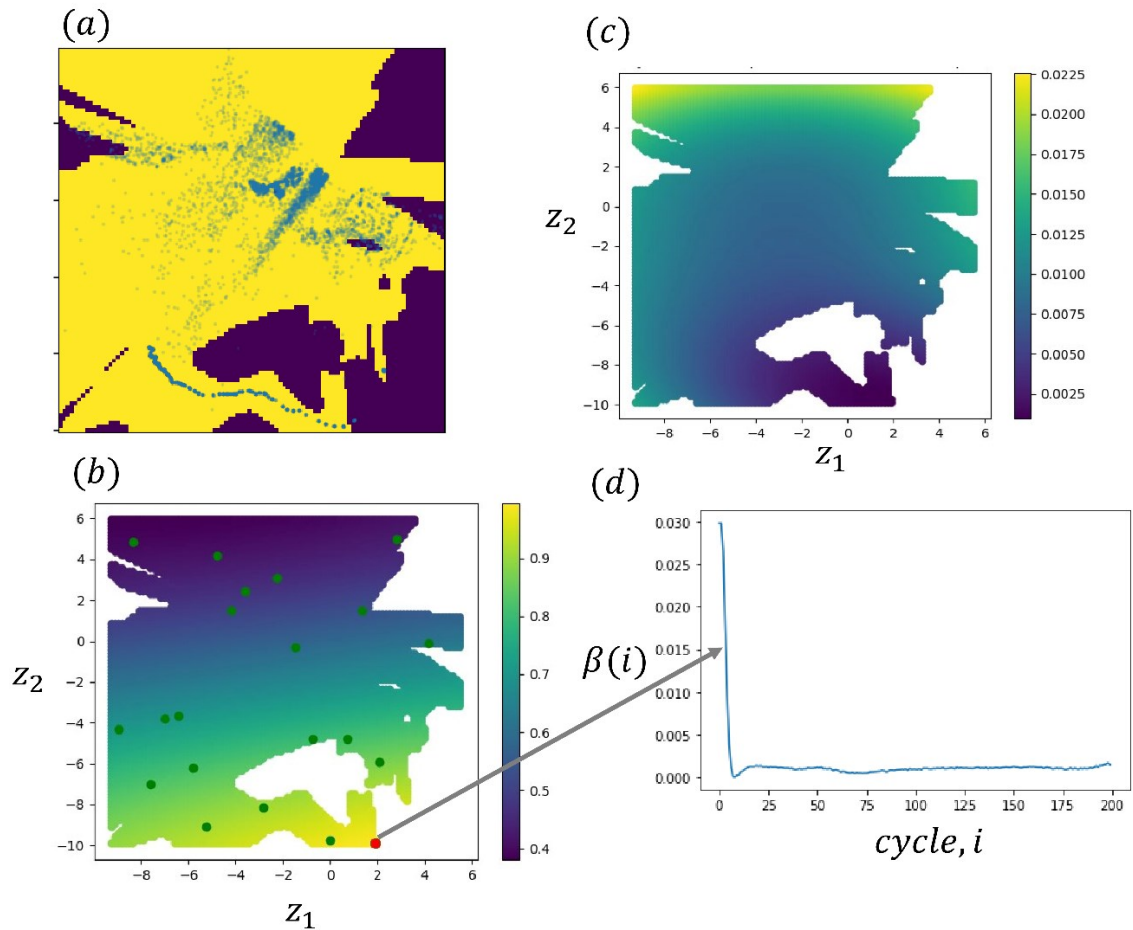

**Figure 10.** Classification problem of Plasmonic Nanoparticles data (7 discrete labels). Results of zBO to find optimal region of $\boldsymbol{\beta_c}$ trajectories. (a) 2D latent representation (through VAE) of the

ᵃsergei2@utk.edu

$\boldsymbol{\beta_c}$ trajectories where the blue dots are the training data (b), (c) GP estimated objective function (SSIM loss) map and its uncertainty map respectively. The red dot is the maximum (optimal) estimated objective value and (e) is the respective reconstructed $\boldsymbol{\beta_c}$ trajectory at that optimal location (red dot) in the feasible 2D latent space.

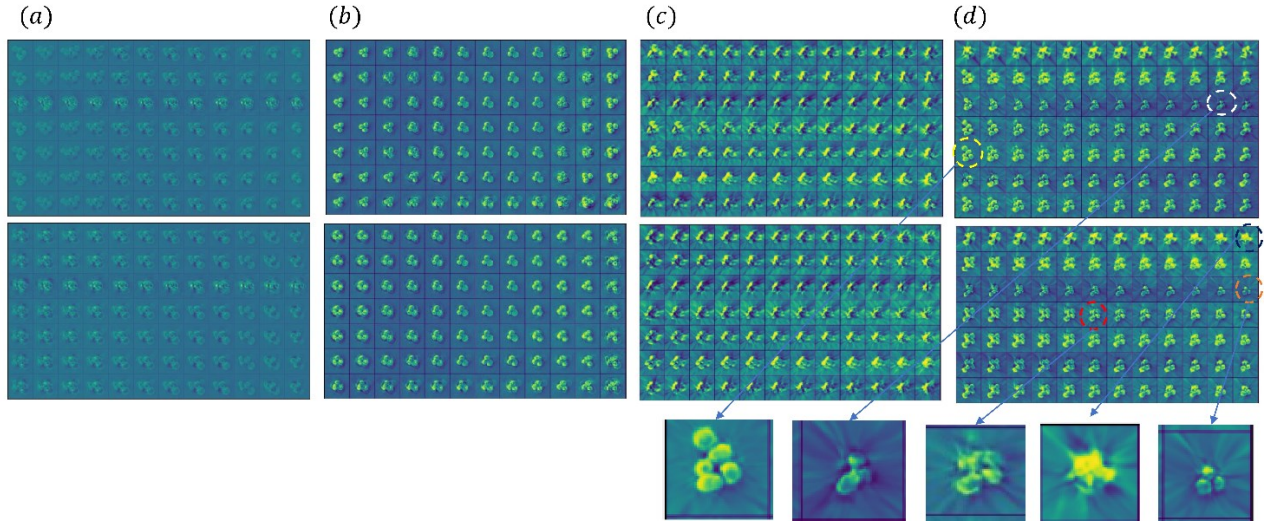

**Figure 11.** Comparison of different VAE models, for multilabel classification of plasmonic nanoparticles dataset, considering 7 discrete labels (particle counts). We have trained all VAE models for 200 cycles. The learned manifolds traversal for the 2D latent space (two latent variables) as shown in rows after training of respective (a) Vanilla VAE model (all parameter in default settings as in [40]), (b) VAE model with other hyperparameter setting as stated for this case study (c) jrVAE model, with default setting of $\boldsymbol{\beta_c}$ trajectory (no zBO implementation) and (d) jrVAE model, with optimal tuning of $\boldsymbol{\beta_c}$ trajectory (zBO implementation). We find 5 discrete labels (as particle counts from 1 to 5) out of 7 labels from the learned manifolds as marked by black, white, orange, red and yellow circles respectively.

## 4. Conclusion

To summarize, here, we extend Latent Bayesian optimization (zBO) to tackle any high-dimensional hyperparameter optimization problem for joint rotationally invariant variational autoencoders. In high dimensional parameter optimization, considering expensive function evaluations, the guidance of where to invest more on the search space is very critical in terms of reducing overall computational cost, thus manual or exhaustive search is very tedious or infeasible. Due to the curse of dimensionality, the standard BO is also not entirely reliable or computationally efficient. In the zBO framework, the high dimensional parameter space is compressed into a two-dimensional latent space, where we capture sufficient variability of parameters through training a variational autoencoder model. We see the latent space preserves the pattern of original training samples (as per domain expert knowledge) while introducing some variability (new hybrid patterns) as well in the feasible input set of solutions. Thus, the prior expert knowledge is still preserved as we project into the reduced latent space. Then, the reduced latent space is considered as the proxy search space for optimization where the optimal latent solution can be easily decoded into the high dimensional trajectory.

[a]sergei2@utk.edu

In this analysis, we choose to optimize the high dimensional continuous scale parameters of joint rotational variational autoencoder (jrVAE) model and apply to multi-label classification problems of MNIST and plasmonic nanoparticles dataset. We see the performance of jrVAE model, with optimal tuning of the scale parameters through zBO is promising. Interestingly, the optimal tuning of scale parameters varies even in patterns for given problems, where an optimal trajectory pattern (eg. Cooldown trajectory) in one problem (MNIST dataset) seemed not ideal for the other problem (plasmonic nanoparticles), and thus cannot be generalized to guarantee best performance of the same model to extract information from all type of dataset. This observation furthers value the need of zBO framework in enhancing a ML model, without the assumption of generalizing optimal tuning, for different problems separately. For any further improvement in our analysis, the framework can be extended to optimize both the high dimensional discrete and continuous scale factors jointly, which is a scope for future. However, the overall approach is flexible to incorporate various pattern of trajectories (from different functionals) in the same latent space, to handle any dimension of input parameters without increasing the dimension of the reduced latent space, to consider different problem objectives (other than classifications) as set my user defined objective functions in the zBO.

**Supplementary Material:**

See the supplementary material for detailed analysis (additional figures) of zBO run, different jrVAE training on Plasmonic nanoparticles datasets.

**Acknowledgements:**

This work was supported by the US Department of Energy, Office of Science, Office of Basic Energy Sciences, as part of the Energy Frontier Research Centers program: CSSAS—The Center for the Science of Synthesis Across Scales—under Award No.DE-SC0019288, located at

[a]sergei2@utk.edu

University of Washington, DC. The autoencoder research was supported by the Center for Nanophase Materials Sciences (CNMS), which is a US Department of Energy, Office of Science User Facility at Oak Ridge National Laboratory. The experimental dataset used in analysis was supported from Ginger Lab, University of Washington. We also thank José Miguel Hernández-Lobato for valuable feedback.

**Conflict of Interest:**

The authors declare no conflict of interest.

**Data Availability Statement:**

The analysis reported here is summarized in Colab Notebook for the purpose of tutorial and application to other models (https://github.com/arpanbiswas52/PaperNotebooks).

[a]sergei2@utk.edu

[a]sergei2@utk.edu

[a]sergei2@utk.edu

[a]sergei2@utk.edu

[a]sergei2@utk.edu

Supplementary Materials of the paper titled

# "Optimizing Training Trajectories in Variational Autoencoders via Latent Bayesian Optimization Approach"

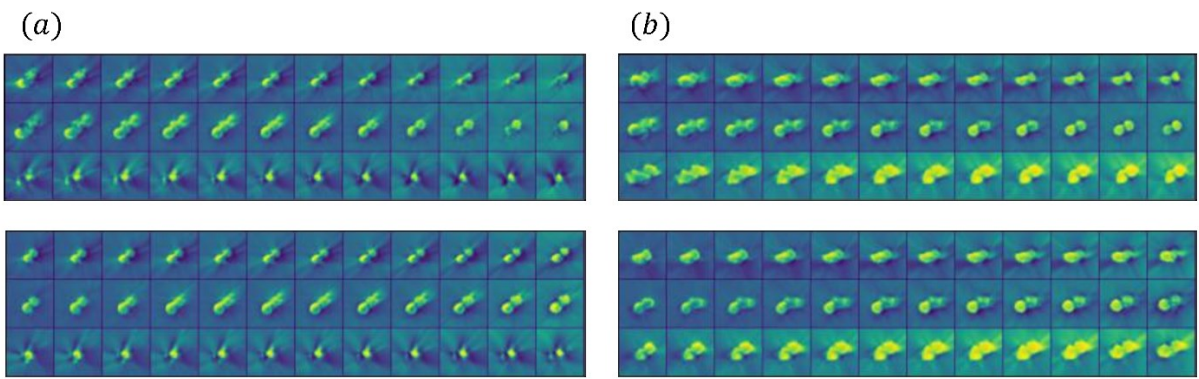

**Figure A1.** Additional comparison of jrVAE models, solving for classification problem of Plasmonic Nanoparticles data (3 discrete labels). We have trained different VAE model for 200 cycles. The learned manifolds traversal for the 2D latent space (two latent variables) as shown in rows after training of respective (a) jrVAE model, with optimal tuning (as shown in fig. 8c) of $\beta_c$ trajectory (zBO implementation), (c) jrVAE model, with similar tuning (as shown in fig. 6c) of $\beta_c$ trajectory optimized in MNIST analysis. In this case, we train the model for 200 cycles where the last 80 cycles of $\beta_c$ were constant with value of the 120$^{th}$ cycle.

Here we see, as stated, the top and the bottom row of the learned manifolds for optimal tuned jrVAE (fig A1a.) shows the discrete labels for two and one particles, where few locations of the middle row able to detect the three-particles class. However, with the optimal scale factors found for MNIST dataset, the performance degrades (not learning the entire physical information) for nanoparticles dataset, where we get two-particles class at most grid locations of the learned manifolds. This shows $\beta_c$ trajectory is sensitive to the problem complexity (dataset) as well and cannot be generalized. This adds value to the need of the zBO approach in finding this optimal tuning of high dimensional parameters, as appropriate for a specific task of a given application.

31[a]sergei2@utk.edu

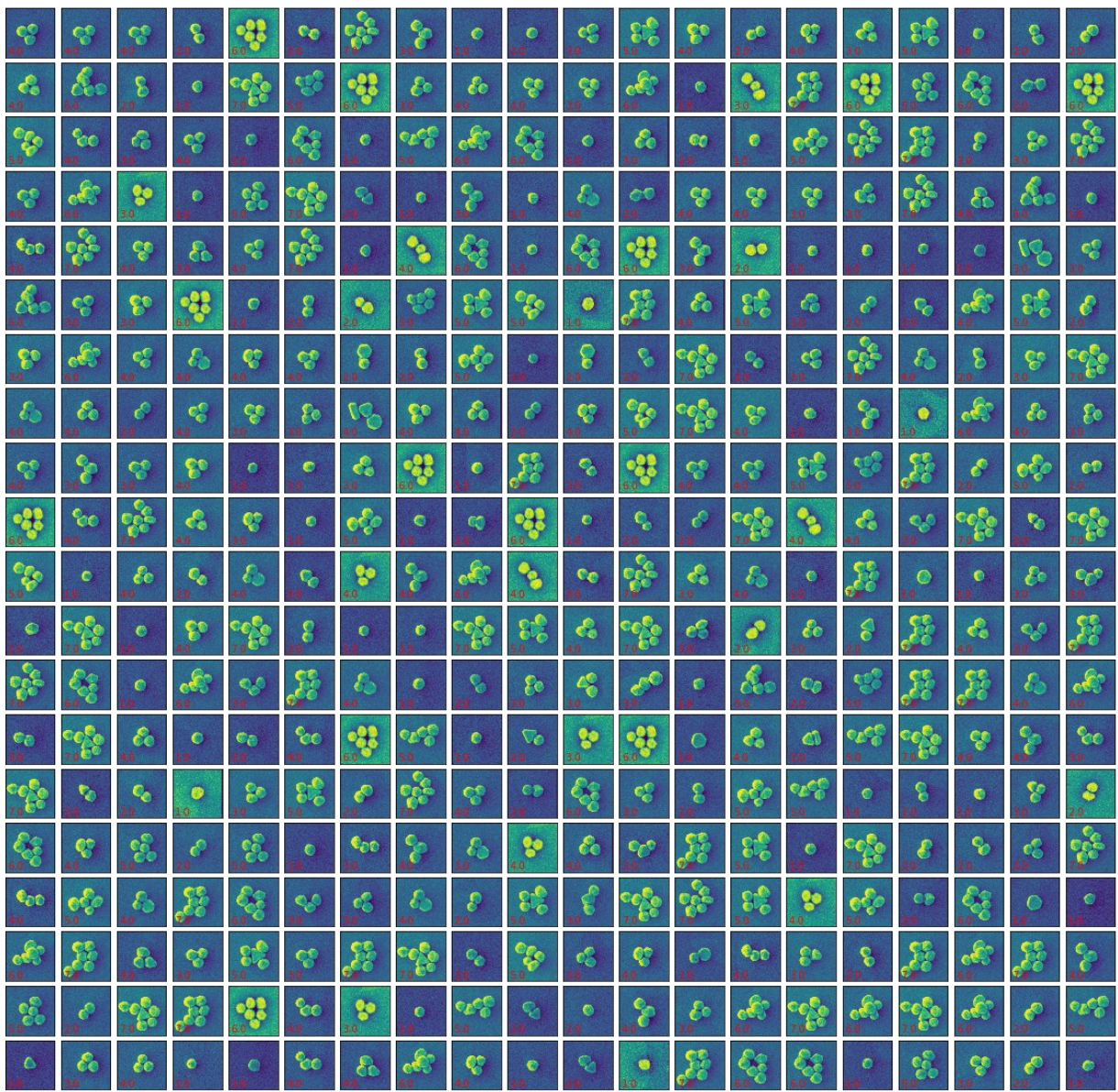

**Figure A2.** Few training samples of plasmonic nanoparticles images, considering 7 discrete labels as particle counts.

[a]sergei2@utk.edu